\documentclass[10pt,a4paper,conference]{IEEEtran}
\usepackage{cite}

\ifCLASSINFOpdf
\usepackage[pdftex]{graphicx}
\else
\fi
\usepackage{amsmath}
\usepackage{algorithmic}

\usepackage{array}

\ifCLASSOPTIONcompsoc
  \usepackage[caption=false,font=normalsize,labelfont=sf,textfont=sf]{subfig}
\else
  \usepackage[caption=false,font=footnotesize]{subfig}
\fi
 \usepackage{dblfloatfix}

\usepackage{url}

\usepackage{color}
\newcommand{\TODO}[1]{{\color{black}#1}}
\begin{document}
%
\title{Superpixel-based Refinement for\\ Object Proposal Generation}

\author{\IEEEauthorblockN{Christian Wilms}
\IEEEauthorblockA{Computer Vision Group\\
University of Hamburg\\
Hamburg, Germany\\
Email: wilms@informatik.uni-hamburg.de}
\and
 \IEEEauthorblockN{Simone Frintrop}
\IEEEauthorblockA{Computer Vision Group\\
University of Hamburg\\
Hamburg, Germany\\
Email: frintrop@informatik.uni-hamburg.de}
}


%


\maketitle

\begin{abstract}
Precise segmentation of objects is an important problem in tasks like class-agnostic object proposal generation or instance segmentation. Deep learning-based systems usually generate segmentations of objects based on coarse feature maps, due to the inherent downsampling in CNNs. This leads to segmentation boundaries not adhering well to the object boundaries in the image. To tackle this problem, we introduce a new superpixel-based refinement approach\footnote{Code is available at: \url{https://www.inf.uni-hamburg.de/spxrefinement}} on top of the state-of-the-art object proposal system AttentionMask. The refinement utilizes superpixel pooling for feature extraction and a novel superpixel classifier to determine if a high precision superpixel belongs to an object or not. Our experiments show an improvement of up to $26.0\%$ in terms of average recall compared to original AttentionMask.
 Furthermore, qualitative and quantitative analyses of the segmentations reveal significant improvements in terms of boundary adherence for the proposed refinement compared to various deep learning-based state-of-the-art object proposal generation systems.
\end{abstract}



%
\IEEEpeerreviewmaketitle

\section{Introduction}
\label{sec:introduction}

Object proposal generation, the class-agnostic generation of object candidates in images, is a vital part of many modern object detection or instance segmentation systems to reduce the search space~\cite{girshick2015fast,ren2015faster,he2017mask,chen2018masklab}. Depending on the task, object candidates are either bounding boxes~\cite{gidaris2016attendbmvc,li2017zoom} or pixel-precise segmentation masks~\cite{Pinheiro2015-deepmask,Pinheiro2016-sharpmask,Hu2017-fastmask,WilmsFrintropACCV2018}. In contrast to instance segmentation, object proposal generation is class-agnostic and thus not limited to the classes seen in training. This leads to a more general approach for discovering objects in images~\cite{Osep19ICRA}.

\begin{figure}[t]
\centering
\subfloat[without our refinement]{\includegraphics[width=0.23\textwidth]{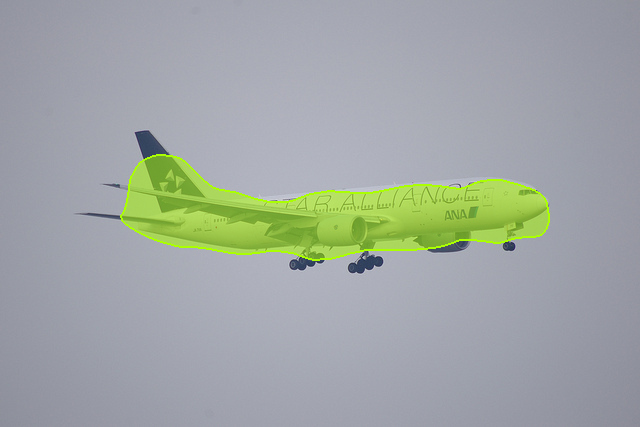}%
\label{fig:comparePx}}
\hfil
\subfloat[with our refinement]{\includegraphics[width=0.23\textwidth]{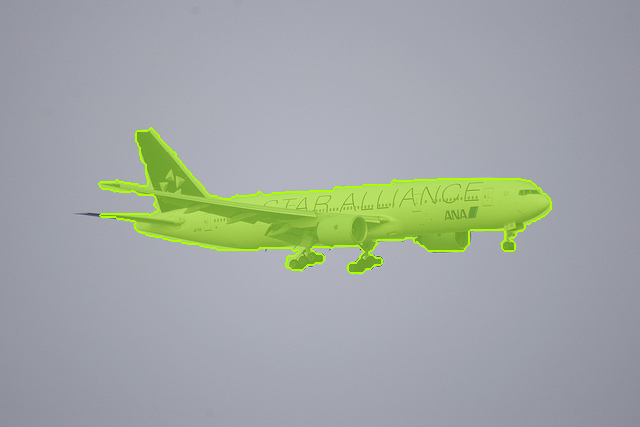}%
\label{fig:comapreSpx}}
\caption{Example of an object proposal without (a) and with (b) refined contours after applying our superpixel-based refinement. Note the precise segmentation of fine details after applying our refinement.}
\label{fig:comparePxSpx}
\end{figure}

Since the advent of deep learning, object proposal generation made big strides~\cite{gidaris2016attendbmvc,li2017zoom,Pinheiro2015-deepmask,Pinheiro2016-sharpmask,Hu2017-fastmask,WilmsFrintropACCV2018}. 
However, deep learning systems easily miss fine details of objects. This is due to the nature of convolutional neural networks (CNNs) subsampling the input to generate semantically rich features. 
While usually not relevant in classification, the loss of fine details is very important for tasks like pixel-precise object proposal generation. Here, detailed contours like the airplane tail or the wheels in Fig.~\ref{fig:comparePxSpx} are usually lost. This is especially a problem for larger objects as those suffer most from downsampling effects~\cite{Hu2017-fastmask,WilmsFrintropACCV2018}.

To tackle this problem, multiple approaches were presented in semantic segmentation or salient object detection. 
Applying a conditional random field (CRF) as post-processing~\cite{hou2017deeply,lin2017refinenet,li2018referring} is a simple solution. Similarly, an encoder-decoder architecture can upsample the coarse results utilizing guidance from earlier feature maps~\cite{li2016deep,chen2017deeplab,lin2017refinenet,li2018referring}. Those are well-established techniques in salient object detection~\cite{hou2017deeply, li2016deep} and semantic segmentation~\cite{chen2017deeplab,lin2017refinenet,li2018referring}. However, object proposal generation is different as individual segmentations are created for the object proposals, which potentially overlap. A CRF would have to be applied to every proposal and an encoder-decoder architecture is only applicable by taking crops of the input image. Thus, both options are computationally demanding \cite{chen2018reverse,Pinheiro2016-sharpmask}.

Other lines of work include dilated convolutions to generate semantically rich features without downsampling~\cite{chen2017deeplab,yu2015multi}. \cite{zhang2019canet,kirillov2019pointrend}~propose iterative adaptation of initial segmentations. These techniques are mostly computationally demanding as well and thus not applicable to hundreds of object proposals generated by object proposal generation systems. Hence, a refinement system is necessary that effectively captures details of objects and shares computation between proposals to be efficient enough for application to hundreds of object proposals.

In classic computer vision, superpixels~\cite{ren2003learning} are used to reduce the number of basic entities in images. Subsequently, this increases the efficiency, while retaining the relevant structures of images. Superpixels as a result of an oversegmentation represent coherent regions of an image. Due to the non-lattice structure of superpixel segmentations, it is difficult to integrate superpixels naturally into CNNs~\cite{he2015supercnn}. However,~\cite{he2017std2p,kwak2017weakly,park2017superpixel} already successfully applied superpixels in CNNs to aggregate information across parts of feature maps.

\begin{figure}[t]
\centering
\includegraphics[width=0.5\textwidth]{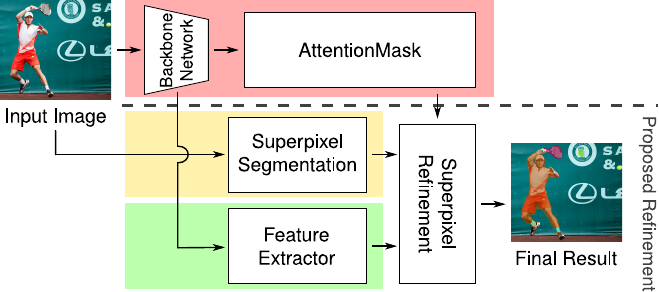}%
\caption{Overview of the proposed superpixel-based refinement on top of AttentionMask~\cite{WilmsFrintropACCV2018}. Superpixel segmentations are generated and features are extracted from the backbone network. The superpixel refinement creates feature vectors per superpixels based on the segmentations and a feature extractor. The per-superpixel feature vectors are combined with the coarse results of AttentionMask and classified for final, more precise object proposals.}
\label{fig:abstractSysFig}
\end{figure}

In this paper, we therefore propose an end-to-end learned superpixel-based refinement approach for object proposal generation systems. As visualized in Fig.~\ref{fig:abstractSysFig}, we utilize the coarse proposal masks of the state-of-the-art object proposal generation system AttentionMask~\cite{WilmsFrintropACCV2018} and the highly accurate superpixels~\cite{felzenszwalb2004efficient}. To combine both, we first apply superpixel pooling. Inspired by~\cite{he2017std2p,kwak2017weakly,park2017superpixel}, we extract per-superpixel features from a backbone network that are shared between proposals. Second, we sample superpixels in and around the coarse object proposal masks for refinement. Finally, we propose a new superpixel classifier to distinguish between superpixels as part of an object or the background. The classifier decides based on the pooled features and the coarse object proposal masks roughly capturing objects. This leads to highly precise results utilizing deep features. Quantitative results on the LVIS dataset~\cite{gupta2019lvis}, which has more precise annotations compared to the MS COCO dataset~\cite{lin2014microsoft}, indicate the effectiveness of our superpixel-based refinement. Additionally, qualitative and quantitative results show a significantly improved boundary adherence compared to state-of-the-art systems. 

Our main contributions are:
\begin{itemize}
\item Utilizing superpixel pooling in a shared and end-to-end learned manner to extract rich features while maintaining highly accurate segmentations capturing fine details.
\item A superpixel classifier to distinguish foreground and background superpixels based on pooled features and a prior derived from coarse object proposals as guidance.
\item Introducing a superpixel-based refinement approach that can be easily integrated into object proposal generation systems 
 to increase the results by up to $26.0\%$.
\end{itemize}

\section{Related Work}
In this section, we give an overview of related work in the areas of object proposal generation and superpixels in CNNs.

\subsubsection{Object Proposal Generation}
\label{sec:relWorkOPG}
Object proposal generation as a new task was first introduced by~\cite{alexe2010object}. 
\cite{Hosang2015PAMI}~give a survey of systems not using deep learning. In general, systems can be distinguished by the type of proposals they generate. The proposals are either bounding boxes \cite{Alexe12PAMI,zitnick2014edgeboxes,gidaris2016attendbmvc,li2017zoom} or pixel-precise segmentation masks \cite{pont2017multiscale,Manen13ICCV,uijlings2013selective,Pinheiro2015-deepmask,Pinheiro2016-sharpmask,Hu2017-fastmask,WilmsFrintropACCV2018}. Systems generating bounding boxes 
 sample windows across the input image and calculate an objectness score per window. The score describes the likelihood of the window containing an object. Additionally, the bounding box is refined from the initial window. Both, hand-crafted~ \cite{Alexe12PAMI,zitnick2014edgeboxes} and deep learning~\cite{gidaris2016attendbmvc,li2017zoom} systems exist. In contrast to those, we are working on pixel-precise segmentation masks. This is more complex as not only the four parameters of the bounding box have to be regressed but a classification of each pixel or superpixel is necessary.

Pixel-precise object proposal generation systems mainly started from grouping superpixels using hand-crafted features \cite{Manen13ICCV,uijlings2013selective}. 
With the emergence of deep learning, superpixels got abandoned from the systems as their integration into the networks is non-trivial. Thus, most systems switched to a window scoring approach \cite{Pinheiro2015-deepmask,Pinheiro2016-sharpmask,Hu2017-fastmask,WilmsFrintropACCV2018}. DeepMask~\cite{Pinheiro2015-deepmask} and SharpMask~\cite{Pinheiro2016-sharpmask} process windows from an image pyramid and generate a pixel-precise segmentation as well as an objectness score per window. SharpMask in contrast to DeepMask uses an encoder-decoder architecture to refine the initial segmentation.


FastMask~\cite{Hu2017-fastmask} reduces the redundancy by generating a feature pyramid within the network, thus extracting features only once per image. From the feature pyramid, windows are densely sampled and objectness scores as well as pixel-precise segmentations are generated.
 Further increasing the efficiency and improving the detection of small objects, AttentionMask~\cite{WilmsFrintropACCV2018} introduces a notion of attention. 
In AttentionMask, only promising windows are sampled from the feature pyramid to reduce computation and allow an additional pyramid level for small objects.
 AttentionMask is the base for our proposed superpixel-based refinement approach and is described in more detail in Sec.~\ref{sec:attMask}. All four approaches, however, suffer from the downsampling in CNNs leading to coarse segmentations.

\subsubsection{Superpixels in CNNs}
Few approaches utilize superpixels for CNNs as the concepts are non-trivial to combine. This is mainly attributed to the missing lattice structure in superpixel segmentations. \cite{he2015supercnn}~circumvent the missing lattice structure and create a vector per superpixel describing the differences to all other superpixels in the image. Those vectors serve as input to a CNN. 
\cite{gadde2016superpixel}~propose a Gaussian bilateral filtering on superpixels to enforce proximate superpixels to have similar output. In salient object detection, \cite{zhao2015saliency} and \cite{tang2016saliency} generate the saliency prediction per superpixel based on its surrounding. However, different to our approach, they use superpixels only to reduce the number of entities, not for pooling features.




More similar to us, \cite{he2017std2p,kwak2017weakly,park2017superpixel,mostajabi2015feedforward} utilize superpixel pooling to generate feature representations per superpixel. Using superpixel pooling, \cite{he2017std2p} aggregate information of similar superpixels along a sequence of frames, while \cite{kwak2017weakly} capture semantic units in weakly supervised semantic segmentation. 
\cite{park2017superpixel} and \cite{mostajabi2015feedforward} use features from superpixels for classification in semantic segmentation.
However, our approach is different, since we use superpixel pooling to generate refined, more detailed segmentations in object proposal generation. 


\section{Base Framework}
\label{sec:attMask}
As mentioned in Sec.~\ref{sec:introduction}, AttentionMask~\cite{WilmsFrintropACCV2018} is the base for our refinement approach and therefore briefly described here. Similar to FastMask~\cite{Hu2017-fastmask}, in AttentionMask the backbone network processes an image only once yielding a feature pyramid within the network. AttentionMask uses a ResNet-50~\cite{he2016-resnet} as backbone. The feature pyramid consists of 5 or 8 scales ($\mathcal{S}_n$), as depicted in the upper part of Fig.~\ref{fig:detailedSysFig}, color-coded in red. Here, $n$ denotes the downscale factor with respect to the input image and ranges from $8$ to $128$. These scales correspond to different object sizes with $\mathcal{S}_{8}$ capturing small objects while $\mathcal{S}_{128}$ captures large objects. 
Per scale, features are extracted and a scale-specific objectness attention module (SOAM) is introduced. The SOAMs are learned components that highlight for each scale the locations of relevant objects within the respective feature map. Consequentially, false positives and the computational effort are reduced.

AttentionMask samples windows of fixed size from the feature pyramid at locations where the scale-specific objectness attention of the corresponding SOAM is high. 
By sampling fixed sized windows across feature maps of different scales, objects of different sizes are detected. For each sampled window the objectness is determined and the attentional head generates a first rough estimation of the object's location. The localization is refined to a coarse pixel-precise segmentation mask of size $40 \times 40$. Hence, the size of the mask is independent of the object size in the input image. This process is visualized in the upper right corner of Fig.~\ref{fig:detailedSysFig}. Note, that we refer here to the AttentionMask$^{8}_{128}$ model in~\cite{WilmsFrintropACCV2018}. However, all subsequent steps can be carried out in combination with the other models from~\cite{WilmsFrintropACCV2018} as well. 

\section{Method}

\begin{figure*}[t]
\centering
\includegraphics[width=0.99\textwidth]{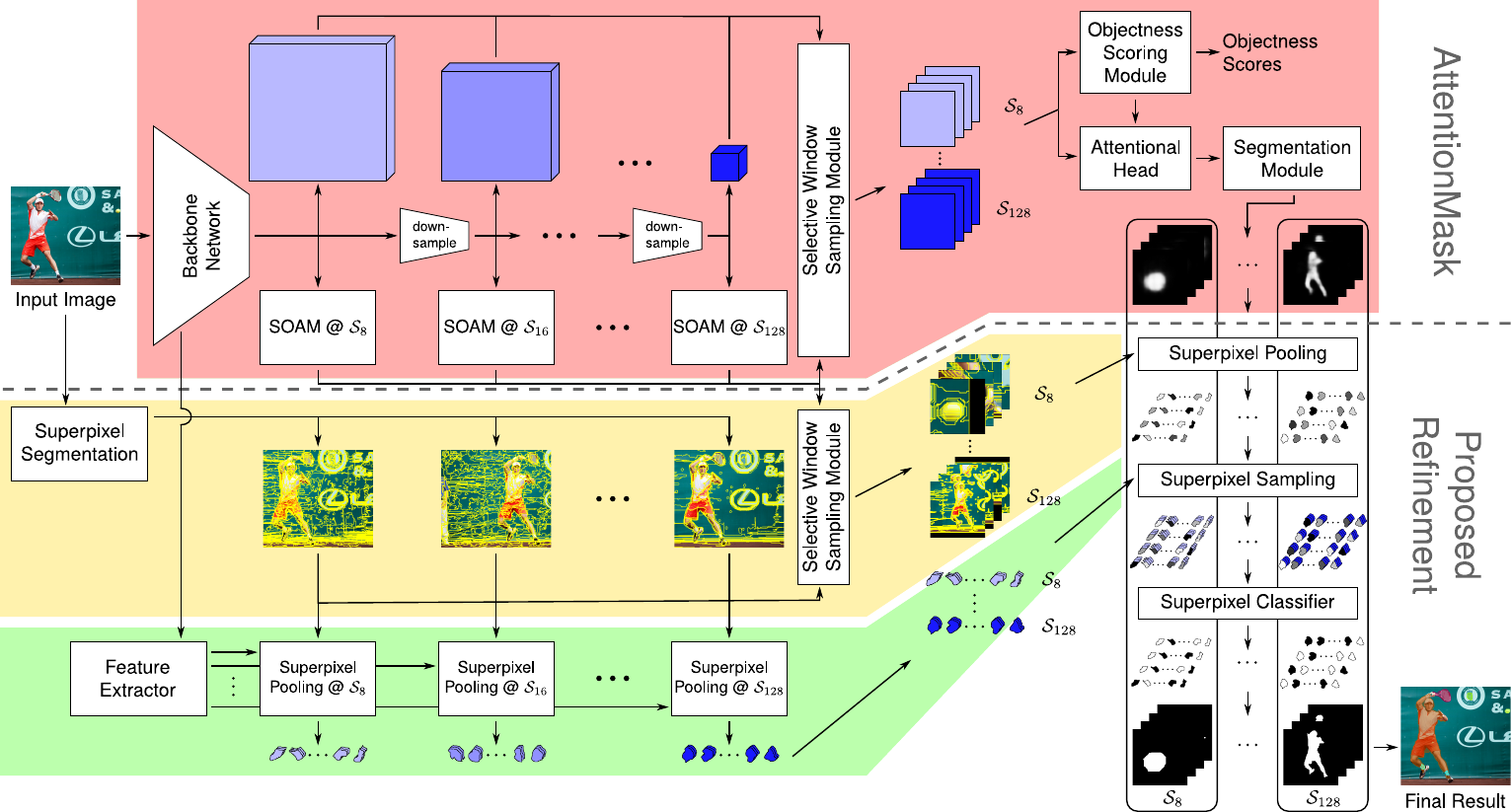}%
\caption{Detailed view of our proposed superpixel-based refinement approach on top of AttentionMask~\cite{WilmsFrintropACCV2018}. We first apply AttentionMask to the input image, generating coarse pixel-precise masks on $40 \times 40$ grids (red). In parallel, a highly accurate superpixel segmentation is generated per scale, which captures fine details of objects (yellow). Based on the segmentations, we apply superpixel pooling to feature maps extracted from the backbone network (green). This leads to per-superpixel feature-vectors. Next, we crop the respective parts of the superpixel segmentations for each AttentionMask result window and apply superpixel pooling on the AttentionMask results. The superpixel sampling combines coarse results and superpixels by concatenating the per-superpixel average of AttentionMask results (gray superpixels) and the per-superpixel features (bluish superpixels). Finally, all superpixels are classified and recombined for high precision results. For clarity in this figure only four windows per scale are considered and the feature extractor generates only three features per superpixel.}
\label{fig:detailedSysFig}
\end{figure*}


In this section, we introduce our novel superpixel-based refinement approach on top of AttentionMask (see Sec.~\ref{sec:attMask}), visualized in Fig.~\ref{fig:detailedSysFig}. 
We start from the final proposals of AttentionMask, which are coarse segmentation masks in $40 \times 40$ windows and apply the refinement per window, i.e. per proposal.
First, the scale $\mathcal{S}_n$ from which each window originates is determined and the according segmentation is chosen. 
Next, we crop the part of the superpixel segmentation spatially corresponding to the window. The cropped segmentation and the AttentionMask result are the input for our superpixel pooling module, described in Sec.~\ref{sec:spxPooling}. The superpixel pooling module calculates the average value of the coarse segmentation mask across every superpixel within the crop, coined mask prior. 
Furthermore, a feature extractor generates per-scale feature maps from the backbone network. Our superpixel pooling module uses these feature maps and creates a feature vector for every superpixel of each scale's segmentation.
To combine these two streams per window, we propose a superpixel sampling module. The superpixel sampling module concatenates for each superpixel of a window the mask prior with the relevant superpixel features. 
Finally, we propose a novel superpixel classifier (see Sec.~\ref{sec:spxCalssifier}) to classify each superpixel based on this information as part of an object or background. The final highly accurate pixel-precise proposals are created using the superpixel segmentations and the classification results. Sec.~\ref{sec:intergration} outlines details of this integration into AttentionMask.

\subsection{Superpixel Pooling and Sampling}
\label{sec:spxPooling}
The first major novelty of the proposed refinement approach is our superpixel pooling module. It enables us to have an output resolution that is largely independent of AttentionMask's final segmentation resolution. Inspired by \cite{he2017std2p,kwak2017weakly,park2017superpixel}, our superpixel pooling is a mixture of global average pooling and standard $n \times n$ average pooling known from CNNs. In superpixel pooling, the superpixels describe the areas for pooling an input feature map. Thus, we pool one feature vector per superpixel in a given segmentation. Similar to the standard pooling, backpropagation through the superpixel pooling module is possible. For this purpose, the gradients are propagated back to the locations of the input feature map corresponding to the superpixels. 


As indicated in Fig.~\ref{fig:detailedSysFig}, we utilize our superpixel pooling module twice in our superpixel-based refinement. First, the module generates per-superpixel features based on feature maps extracted from the backbone net. This is applied once on each scale with a unique segmentation and feature map per scale as visualized in the bottom part of Fig.~\ref{fig:detailedSysFig}, color-coded in green. Second, visualized in the right part of Fig.~\ref{fig:detailedSysFig}, we apply superpixel pooling to each of the AttentionMask result windows using corresponding crops of the superpixel segmentations.
Thus, the windows that are coarsely segmented on a $40 \times 40$ grid by AttentionMask are cropped from the segmentations of the respective scales. 
The central part of Fig.~\ref{fig:detailedSysFig} depicts this process, color coded in yellow. Note that the crops of the segmentations are larger (here: $80 \times 80$ to $1280 \times 1280$) to generate more precise segmentations. The result of the second superpixel pooling step is a superpixelized version of the AttentionMask results
 denoted as mask prior. This prior helps to capture entire objects as~\cite{kwak2017weakly} demonstrate.

Subsequently, our novel superpixel sampling module combines the two streams of superpixel pooling. 
Per window of the AttentionMask results, the superpixel sampling module concatenates the mask prior and the corresponding superpixel features for every superpixel within this window. 
 The lower right part of Fig.~\ref{fig:detailedSysFig} visualizes this concatenation of the mask prior (gray superpixels) and pooled superpixel features (bluish superpixels). Finally, we concatenate the per window results of the sampling across all windows to form one batch of superpixels for classification. Note, that the superpixel pooling of features has to be applied only once per scale and superpixel. However, the superpixel pooling of the mask prior has to be applied to each result of AttentionMask individually.



\subsection{Superpixel Classifier}
\label{sec:spxCalssifier}
\begin{figure*}[t]
\centering
\includegraphics[width=\textwidth]{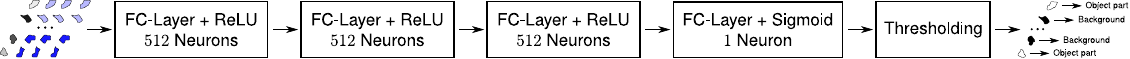}%
\caption{Architecture of our proposed superpixel classifier with fully-connected layers denoted as FC-layer. The input is a batch of superpixels across different windows and scales (variations of blue). For each superpixel, it consists of one mask prior value (gray) from the AttentionMask result and 512 learned features (bluish) extracted from the \textit{res2}-block of the backbone network. As a result the classifier determines if a superpixel is part of an object (white) or part of the background (black).}
\label{fig:spxClassifier}
\end{figure*}

As another major novelty, we propose a superpixel classifier to distinguish between superpixels belonging to objects and background. The classifier, located in the bottom right section of Fig.~\ref{fig:detailedSysFig}, takes the result of the superpixel sampling module, the concatenation of the mask prior and the pooled feature vector per superpixel, as input. The architecture consists of three fully-connected layers with 512 neurons and ReLU activation each as well as a  final fully-connected layer with sigmoid activation (see Fig.~\ref{fig:spxClassifier}). Compared to other structures 
 the proposed architecture is effective, as the results in Tab.~\ref{tab:classifier} reveal.
The classifier processes the superpixels individually without interaction between superpixels.
All superpixels of a scale that do not overlap with a window and therefore are neither sampled nor classified are background for that window.

\begin{table}[t]
\renewcommand{\arraystretch}{1.2}
\centering
\caption{Architectures for the proposed superpixel classifier. \TODO{The first column indicates the amount of neurons per fully-connected layer. AR@$n$ denotes the average recall for the first $n$ proposals.}
}
\label{tab:classifier}
{
\begin{tabular}{@{}lcccccc@{}}
\hline
Architecture &  AR@10 & AR@100  & AR@1000 \\ \hline
$\mathbf{512-512-512}$ & \textbf{0.160} &  \textbf{0.296} & \textbf{0.382} \\
$1024-1024-1024$ & 0.156 &  0.291 & 0.381  \\
$256-256-256$ & 0.158 &  0.292 & 0.380  \\
$512-512-512-512$ & 0.156 &  0.288 & 0.377  \\
$512-512$ & 0.154 &  0.283 & 0.373  \\
\hline
\end{tabular}
}
\end{table}

\subsection{Integration into AttentionMask}
\label{sec:intergration}

To integrate our novel superpixel-based refinement into AttentionMask, we first substitute the backbone network from~\cite{WilmsFrintropACCV2018} for a ResNet-34~\cite{he2016-resnet}. This is necessary as the original model utilizes almost all the GPU memory and thus does not allow for any refinement steps. Changing the backbone leads to only a small degradation as the results in Sec.~\ref{sec:resLVIS} indicate. Following \cite{kirillov2019pointrend}, we utilize the output of the \textit{res2}-block to extract features from the backbone. Results using features from other levels of the backbone (see Tab.~\ref{tab:features}) justify this decision. 
The rest of AttentionMask stays unchanged with all added components learned end-to-end as detailed in Sec.\ref{sec:implDetails}.

\begin{table}[t]
\renewcommand{\arraystretch}{1.2}
\centering
\caption{Results of the proposed refinement approach using features from different ResNet-blocks in feature extraction. \TODO{AR@$n$ denotes the average recall for the first $n$ proposals.}}
\label{tab:features}
{
\begin{tabular}{@{}lcccccc@{}}
\hline
ResNet-block &  AR@10 & AR@100  & AR@1000 \\ \hline
\textit{res1} & 0.158 &  0.292 & 0.377 \\
\textbf{\textit{res2}} & \textbf{0.160} & \textbf{0.296} & \textbf{0.382} \\
\textit{res3} & 0.156 &  0.292 & 0.380 \\
\textit{res4} & 0.153 &  0.287 & 0.372 \\
\hline
\end{tabular}
}
\end{table}

Outside the network we generate superpixel segmentations of the input image with the approach of~\cite{felzenszwalb2004efficient} (FH). FH generates superior results compared to other segmentation approaches (see Sec.~\ref{sec:ablInflSeg}). The segmentations differ in the number of superpixels between scales. As FH does not allow to explicitly set a number of superpixels, we optimize the free parameter in FH. For best performance, we generate roughly 8000 superpixels for $\mathcal{S}_8$ and 500 superpixels for $\mathcal{S}_{128}$ with regular intermediate steps for the intermediate scales.

After superpixel pooling, sampling and classification, we recombine the results per window to form high precision segmentation masks that capture even fine object details. These masks are further post-processed in three steps. First, bilateral filtering~\cite{tomasi1998bilateral} is applied on the level of superpixels using color information. \TODO{Second, we apply morphological opening and closing. Closing removes small holes within superpixels. Opening eliminates segmentation artifacts like thin protuberances extending from the superpixel boundaries, which are frequently occurring in FH segmentations. Finally, we apply non-maximum suppression to the set of proposals with a high intersection over union of $0.95$. This leads to the removal of near duplicates, which are more likely to occur when using superpixels as the number of basic entities is smaller.} 

\section{Implementation Details} 
\label{sec:implDetails}
This section presents technical details of our overall object proposal generation system. In terms of trainable modules, AttentionMask is mainly extended by the novel superpixel classifier (Sec.~\ref{sec:spxCalssifier}). Other components like superpixel pooling or sampling do not learn features as they only restructure the input. Another component, which learns features, is the feature extractor. The feature extractor comprises a $1 \times 1$ convolution on the feature maps extracted from the backbone.

To train the superpixel classifier and in turn the feature extractor as well, we use cross entropy loss as the loss function. We integrate this new superpixel classification loss $\mathcal{L}_{spx}$ into the overall loss function $\mathcal{L}$ of AttentionMask by adding it with a weight factor $w_{spx}$ (here $w_{spx}=1$):

\begin{equation}
\begin{split}
\mathcal{L} = & w_{objn}\mathcal{L}_{objn}+w_{ah}\mathcal{L}_{ah}+w_{seg}\mathcal{L}_{seg} + \\
&\quad w_{att} \sum_n \mathcal{L}_{att_n} + w_{spx} \mathcal{L}_{spx}.
\end{split}
\end{equation}

$\mathcal{L}_{objn}, \mathcal{L}_{ah}, \mathcal{L}_{seg}$, and $\mathcal{L}_{att_n}$ denote the losses from~\cite{WilmsFrintropACCV2018} for the objectness score, the attentional head, the pixel-precise segmentation and the SOAMs of the different scales $\mathcal{S}_n$ with weights respectively. Therefore, the entire system can be trained end-to-end in one step. For training, similar to~\cite{Pinheiro2015-deepmask,Pinheiro2016-sharpmask,Hu2017-fastmask,WilmsFrintropACCV2018}, we use the training set of the MS COCO dataset~\cite{lin2014microsoft}. The backbone network is initialized with ImageNet weights~\cite{he2016-resnet} while the rest of the system is learned from scratch. We use SGD as the optimizer and an initial learning rate of $0.0001$.

Selecting the ground truth for the superpixel classifier is non-trivial since the annotations of the MS COCO dataset~\cite{lin2014microsoft} used for training are pixel-precise. Therefore, some superpixels cover both object and non-object pixels. To solve this, we generate a set of superpixels leading to an optimal intersection over union (IoU) for every annotated object. For this purpose, we start with a superpixel completely contained in the annotation. If this does not hold true for any superpixel, we start from the superpixel with the highest IoU with the annotated object. Subsequently, we greedily add all superpixels that increase the IoU between the annotated object and the set of superpixels. The result is an optimal set of superpixels per annotation.

\begin{table*}[t]
\renewcommand{\arraystretch}{1.2}
\centering
\caption{Results on the LVIS dataset using average recall (AR). AR$^S$, AR$^M$ and AR$^L$denote results on small, medium and large objects.}
\label{tab:resultsLVIS}
\begin{tabular}{@{}lccccccc@{}}
\hline
Method & Backbone &  AR@10 & AR@100  & AR@1000 & AR$^S$@100  & AR$^M$@100  & AR$^L$@100  \\ \hline
MCG~\cite{pont2017multiscale} & - & 0.048  &  0.131 & 0.237  & 0.031  & 0.204  &  0.462                   \\ \hline
DeepMask~\cite{Pinheiro2015-deepmask} & ResNet-50 & 0.069  &  0.147 & 0.214  & 0.014  & 0.314  &  0.430     \\ 
SharpMask~\cite{Pinheiro2016-sharpmask} & ResNet-50 & 0.073  &  0.154 & 0.229  & 0.014  & 0.327  &  0.460     \\ 
FastMask~\cite{Hu2017-fastmask} & ResNet-50 & 0.069 &  0.161 & 0.256  & 0.055  & 0.296 & 0.386                     \\ 
AttentionMask~\cite{WilmsFrintropACCV2018} & ResNet-50 & 0.073 &  0.189 & 0.284  & 0.081  & 0.312  & 0.446                      \\ \hline
AttentionMask~\cite{WilmsFrintropACCV2018} & ResNet-34  & 0.076 &  0.185 & 0.271  & 0.083  & 0.305  & 0.423                      \\
Ours & ResNet-34  & \textbf{0.092} &  \textbf{0.206} & \textbf{0.290} & \textbf{0.092} & \textbf{0.340} & \textbf{0.473}                      \\ 
\hline
\end{tabular}
\end{table*}

\begin{figure*}[t]
\centering
\begin{tabular}{ccccc}
\subfloat{\includegraphics[width = .175\linewidth]{./figs/COCO_val2017LVIS_000000176606.jpg}} &
\subfloat{\includegraphics[width = .175\linewidth]{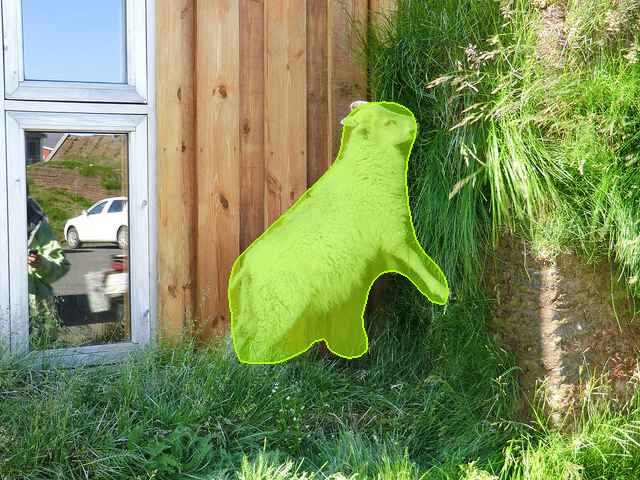}} &
\subfloat{\includegraphics[width = .175\linewidth]{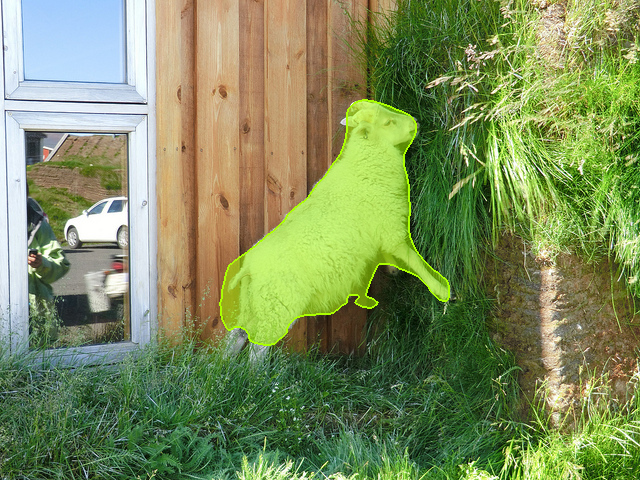}} &
\subfloat{\includegraphics[width = .175\linewidth]{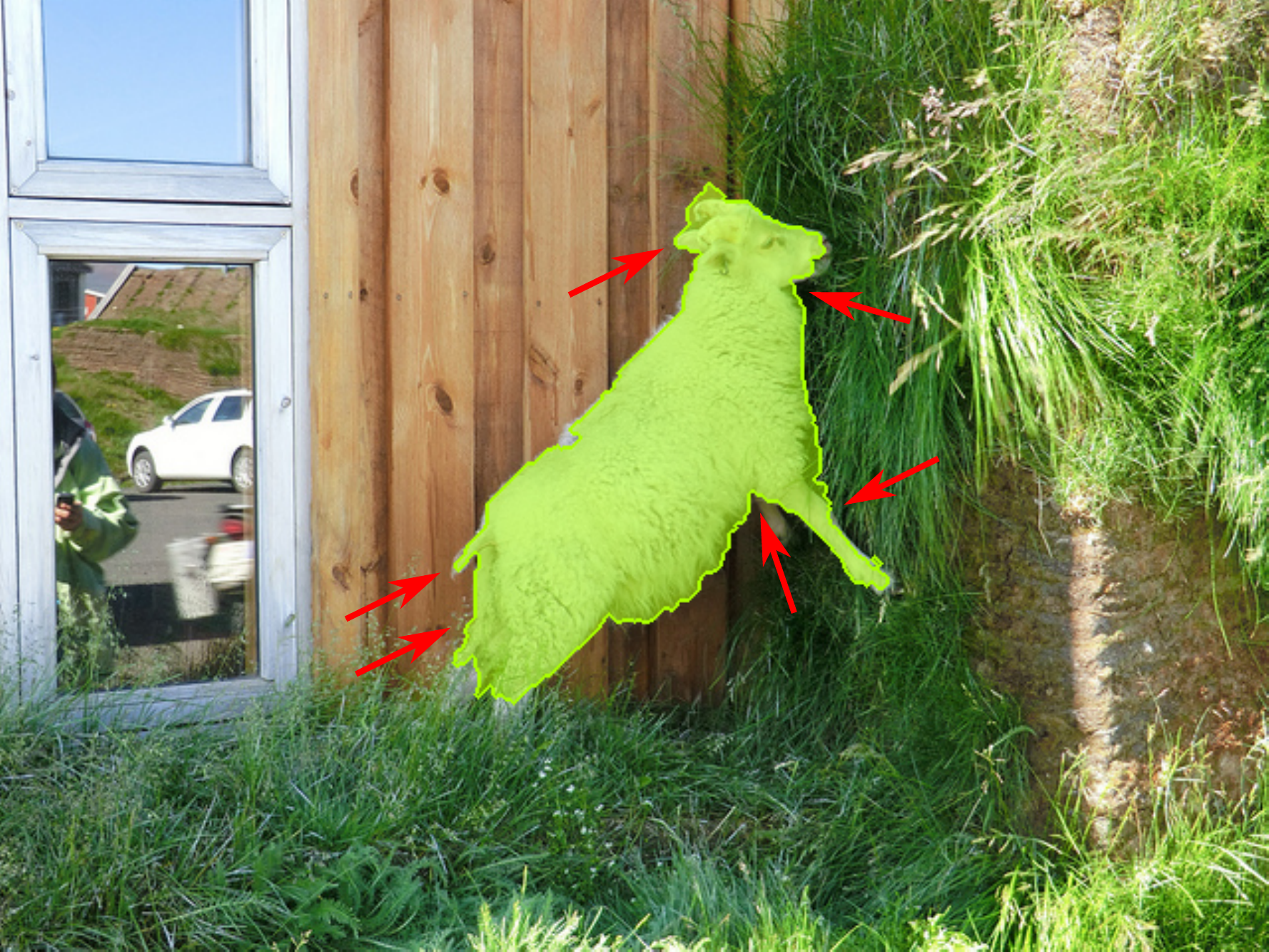}} &
\subfloat{\includegraphics[width = .175\linewidth]{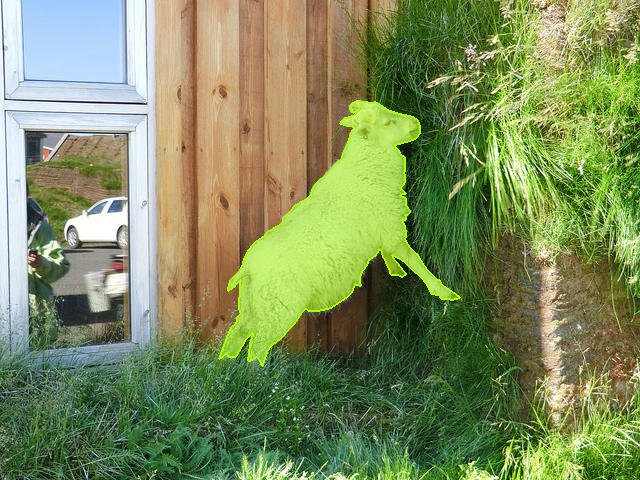}} 
\vspace{-3mm}\\ 
\subfloat{\includegraphics[width = .175\linewidth]{./figs/COCO_val2017LVIS_000000186345.jpg}} &
\subfloat{\includegraphics[width = .175\linewidth]{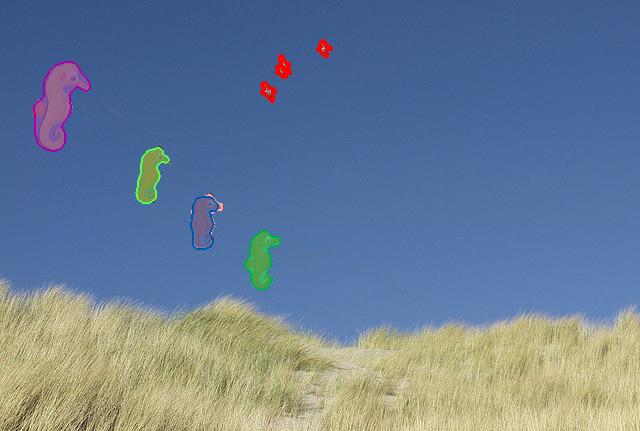}} &
\subfloat{\includegraphics[width = .175\linewidth]{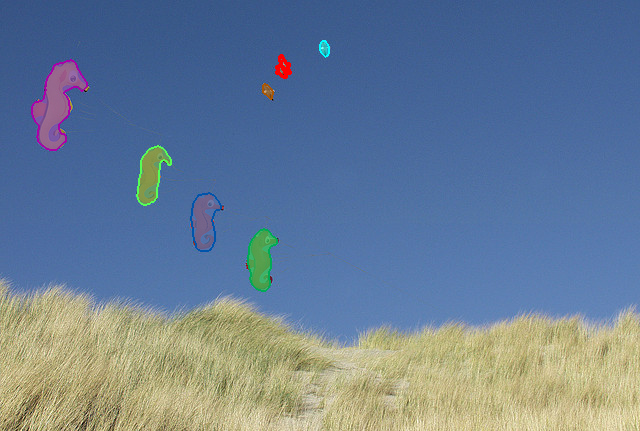}} &
\subfloat{\includegraphics[width = .175\linewidth]{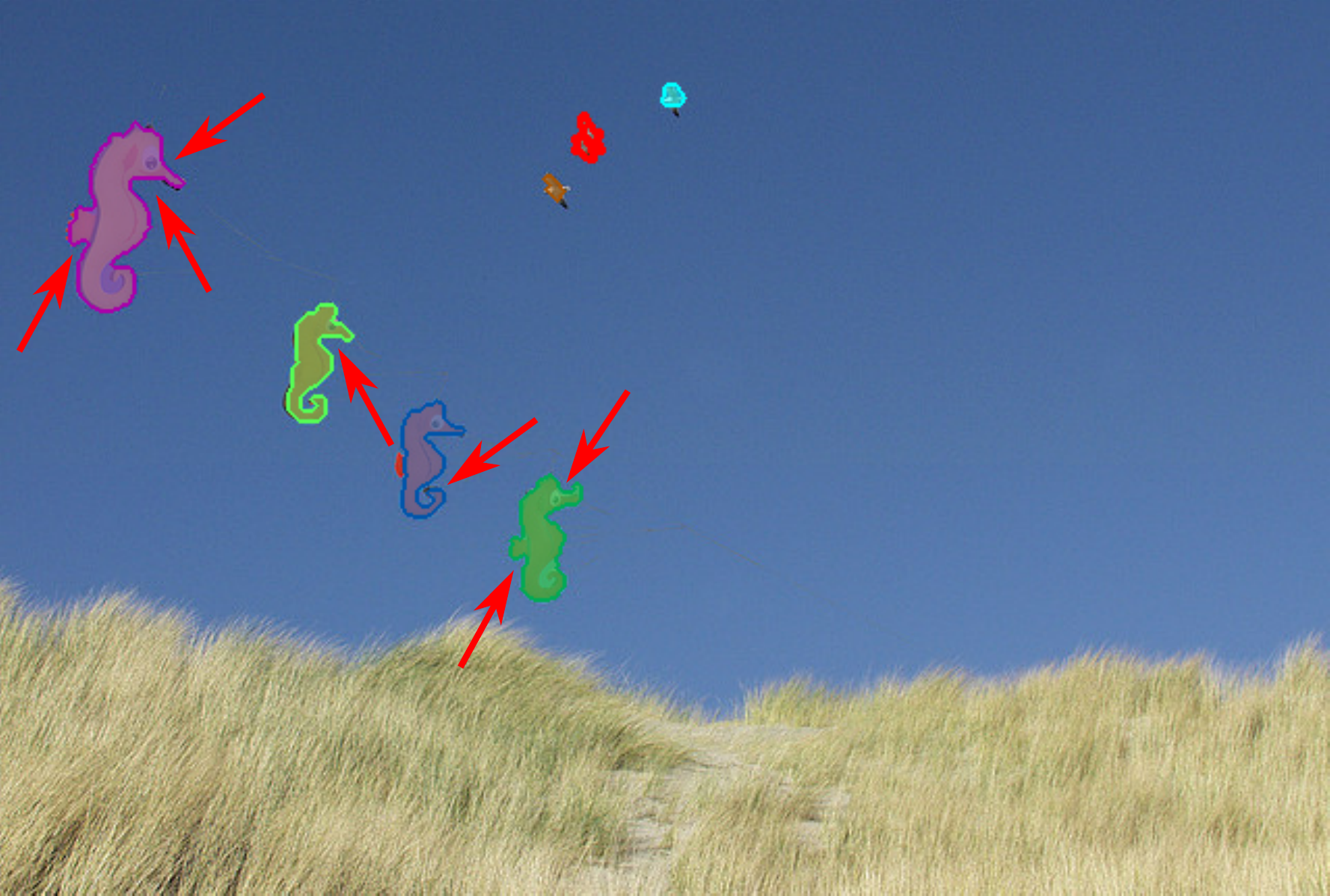}} &
\subfloat{\includegraphics[width = .175\linewidth]{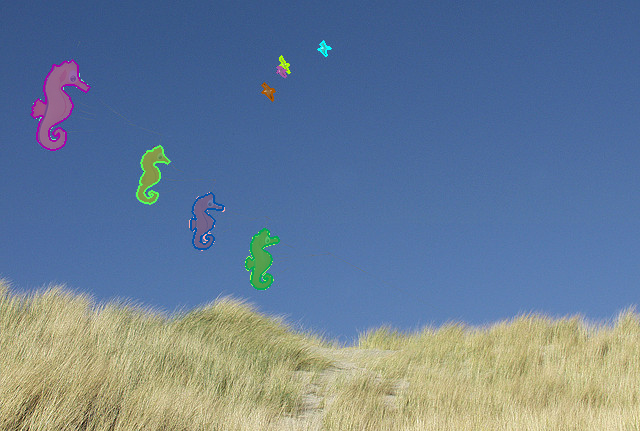}}
\vspace{-3mm}\\
\subfloat{\includegraphics[width = .175\linewidth]{./figs/COCO_val2017LVIS_000000152686.jpg}} &
\subfloat{\includegraphics[width = .175\linewidth]{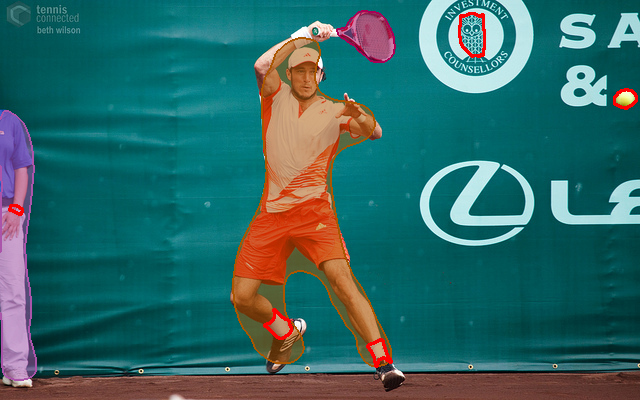}} &
\subfloat{\includegraphics[width = .175\linewidth]{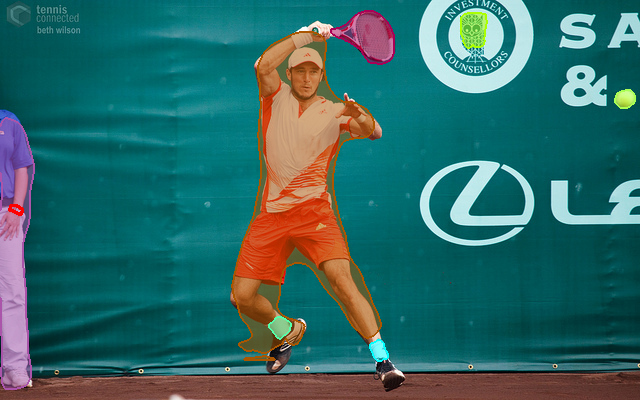}} &
\subfloat{\includegraphics[width = .175\linewidth]{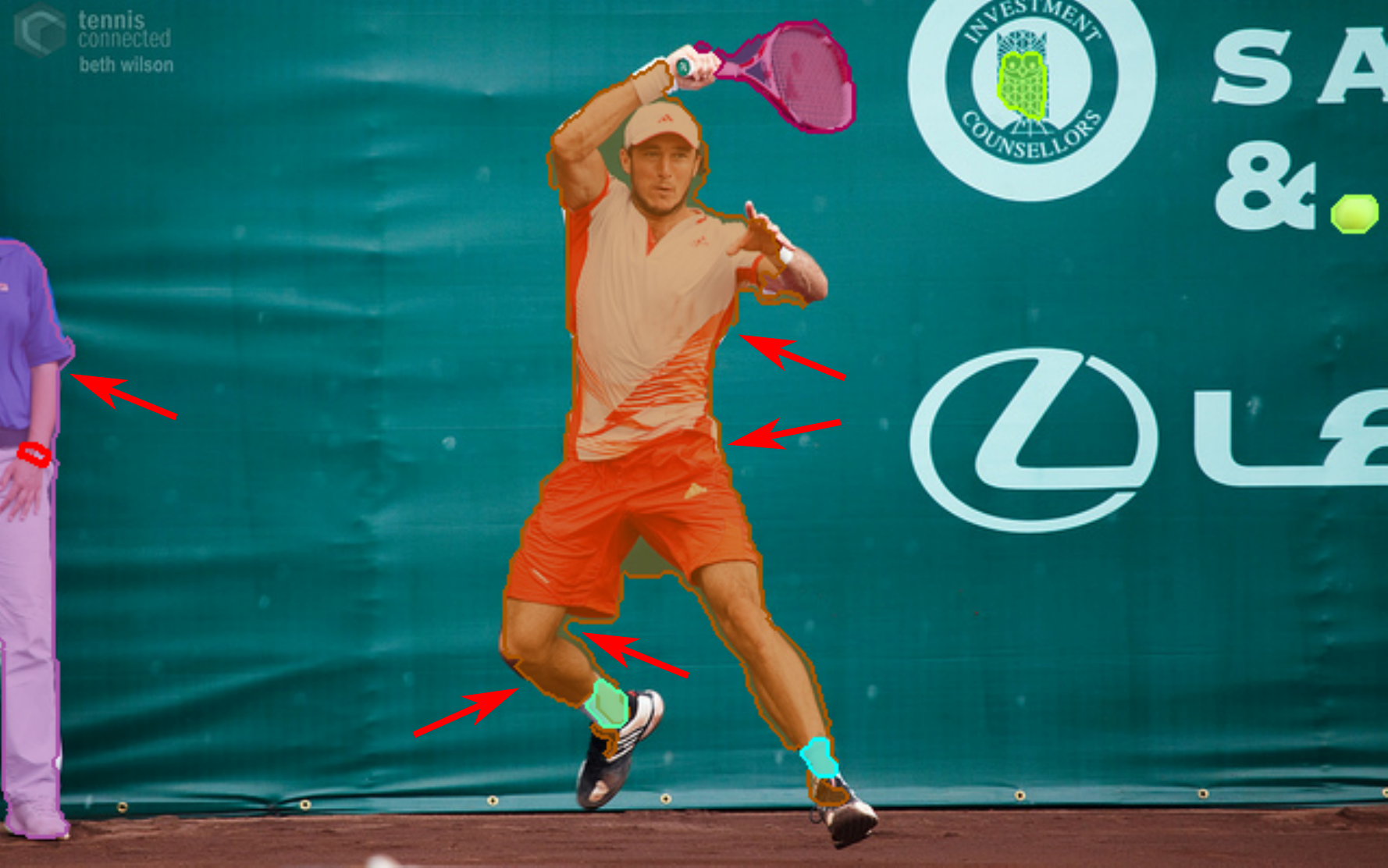}} &
\subfloat{\includegraphics[width = .175\linewidth]{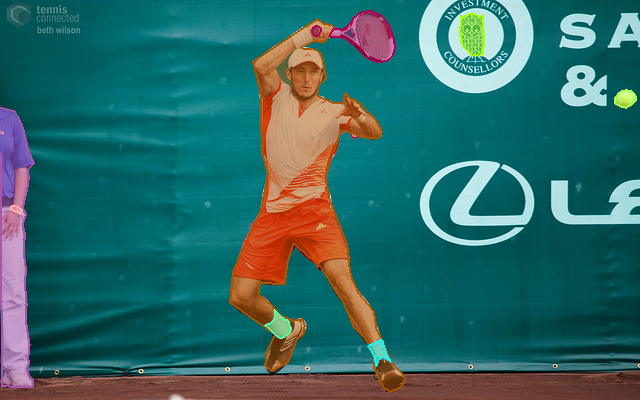}}
\vspace{-3mm}\\
\subfloat{\includegraphics[width = .175\linewidth]{./figs/COCO_val2017LVIS_000000081988.jpg}} &
\subfloat{\includegraphics[width = .175\linewidth]{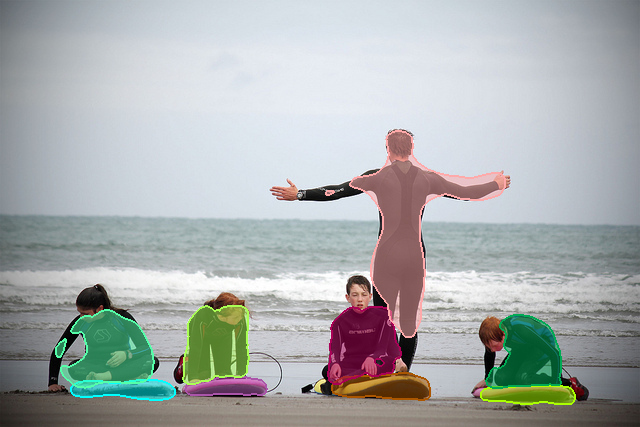}} &
\subfloat{\includegraphics[width = .175\linewidth]{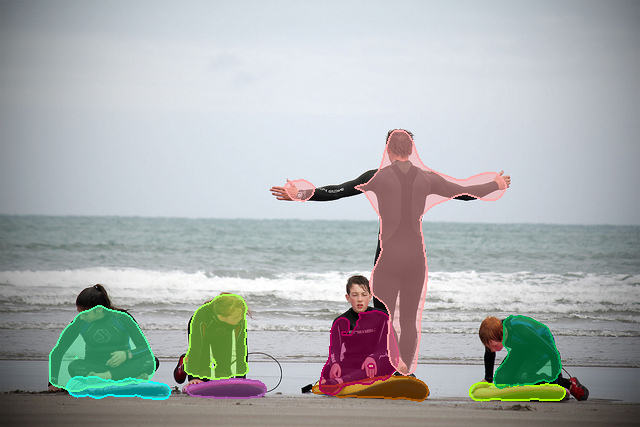}} &
\subfloat{\includegraphics[width = .175\linewidth]{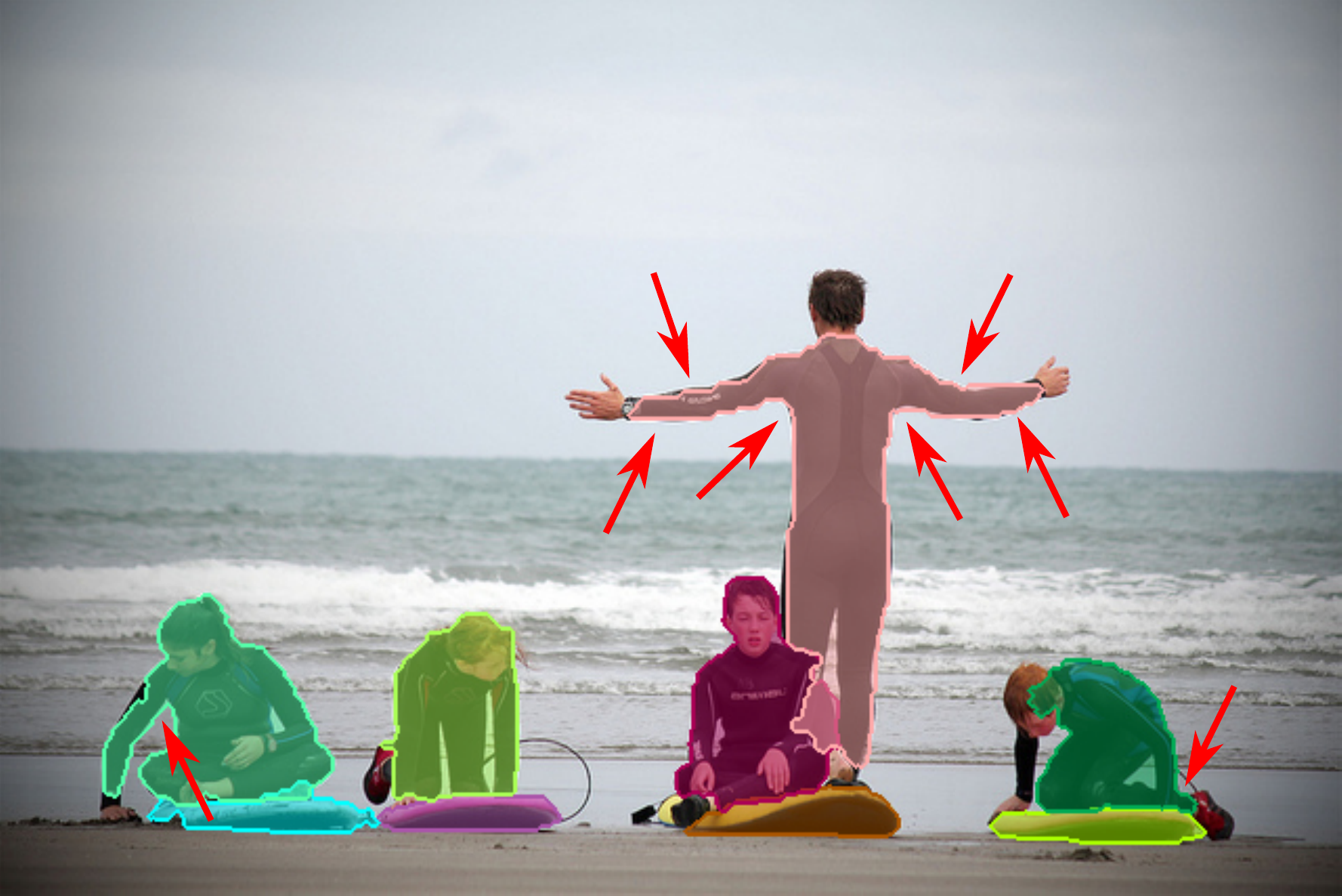}} &
\subfloat{\includegraphics[width = .175\linewidth]{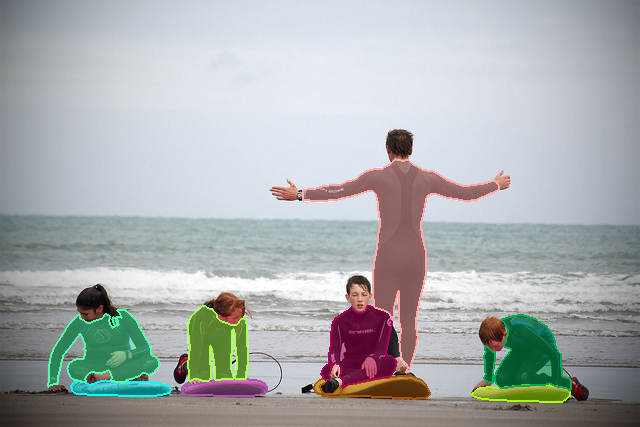}}
\vspace{-3mm}\\
\subfloat{\includegraphics[width = .175\linewidth]{./figs/COCO_val2017LVIS_000000127987.jpg}} &
\subfloat{\includegraphics[width = .175\linewidth]{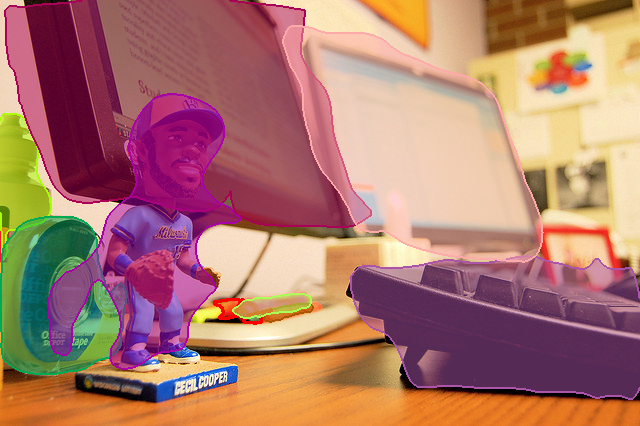}} &
\subfloat{\includegraphics[width = .175\linewidth]{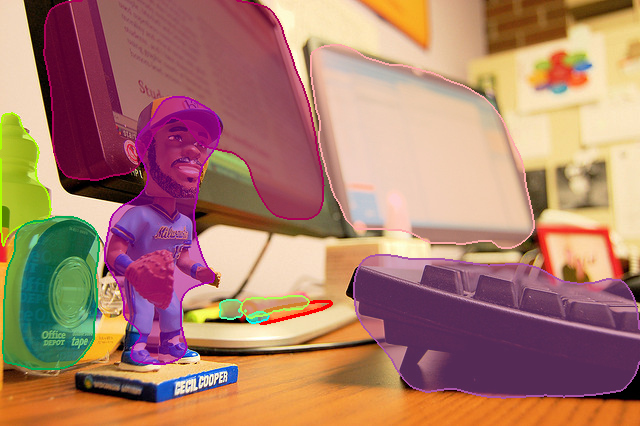}} &
\subfloat{\includegraphics[width = .175\linewidth]{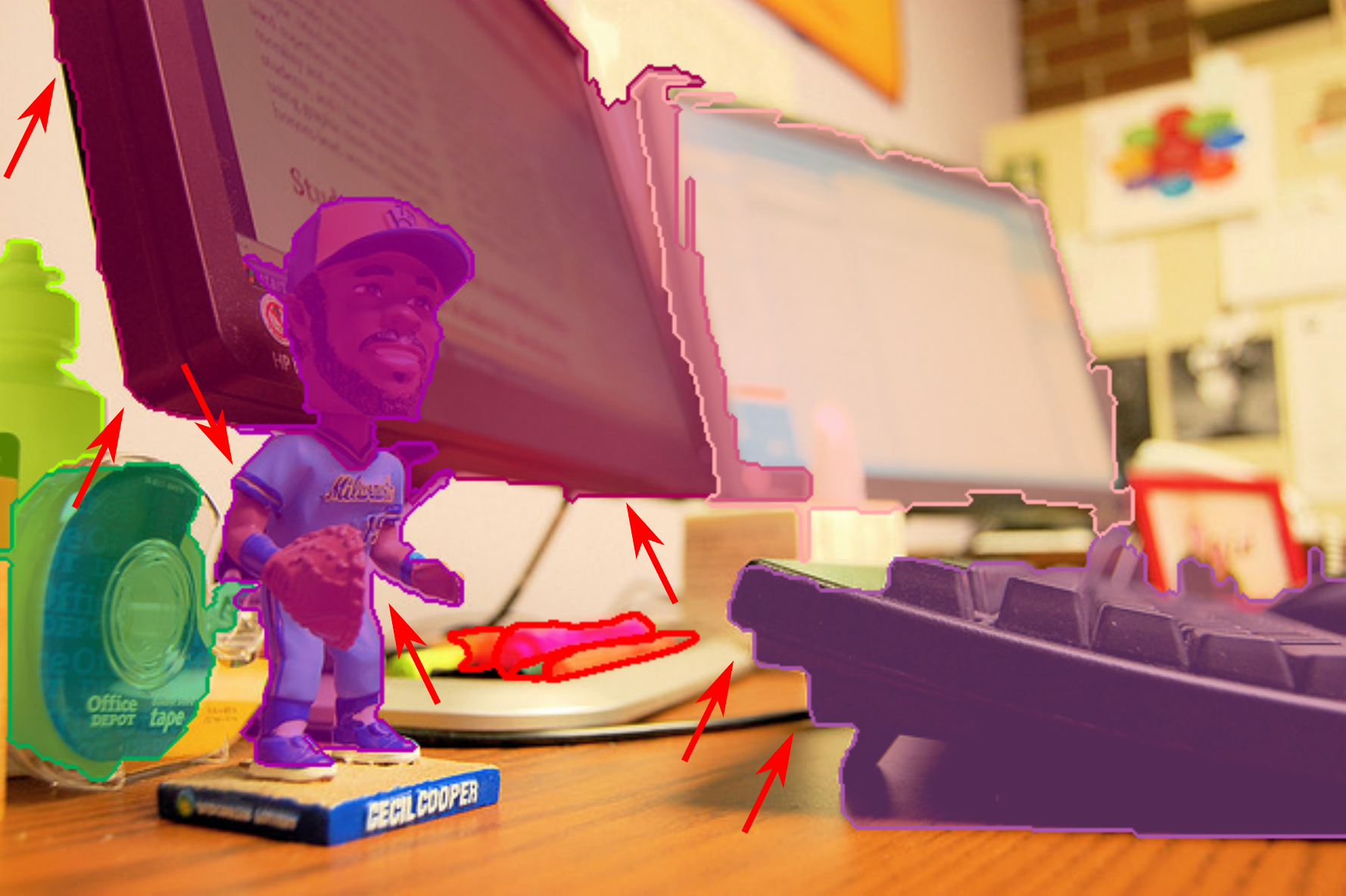}} &
\subfloat{\includegraphics[width = .175\linewidth]{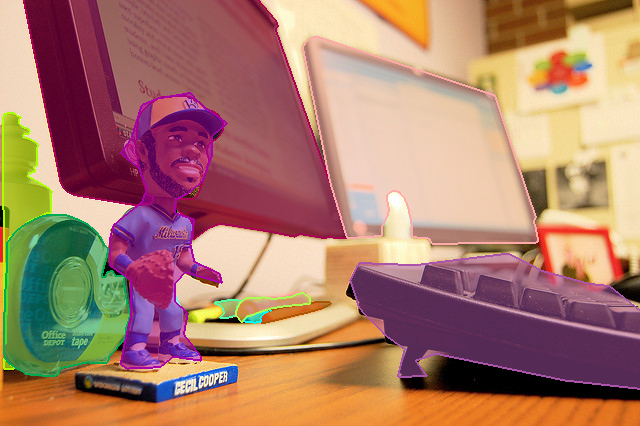}}
\vspace{-1mm} \\ 
{\scriptsize Image}&{\scriptsize FastMask~\cite{Hu2017-fastmask}} &{\scriptsize AttentionMask~\cite{WilmsFrintropACCV2018} (ResNet-50)}&{\scriptsize Ours}&{\scriptsize Ground Truth}
\end{tabular}
\caption{Qualitative results of FastMask~\cite{Hu2017-fastmask}, AttentionMask~\cite{WilmsFrintropACCV2018} and our approach on the LVIS dataset. The filled colored contours denote found objects, while not filled red contours denote missed objects. The red arrows highlight locations showing the strength of our proposals adhering well to object boundaries even in complex scenarios.
}
\label{fig:qualRes}
\end{figure*}


\section{Experiments}
In this section, we present qualitative and quantitative results of our novel superpixel-based refinement approach. 
In contrast to \cite{Pinheiro2015-deepmask,Pinheiro2016-sharpmask,Hu2017-fastmask,WilmsFrintropACCV2018}, we carry out the evaluation on the very recent LVIS dataset~\cite{gupta2019lvis}. The LVIS dataset contains images from the MS COCO dataset's test split~\cite{lin2014microsoft} with more accurate annotations. More accurate annotations are necessary for evaluating our system, as it aims to recover fine details of objects. An evaluation on the MS COCO dataset itself is not conducted. However, \cite{WilmsFrintropACCV2018} already showed the superior performance of AttentionMask compared to other state-of-the-art object proposal systems on the MS COCO dataset.

Following~\cite{Pinheiro2016-sharpmask,Hosang2015PAMI}, the results are reported in terms of average recall (AR) for 10, 100 and 1000 proposals as well as for small, medium and large objects. AR describes how many annotated objects are found and how well those objects are segmented. Note that all results and annotations are pixel-precise masks. 
We compare our system to original AttentionMask~\cite{WilmsFrintropACCV2018} based on two different ResNets as backbone as well as to DeepMask~\cite{Pinheiro2015-deepmask}, SharpMask~\cite{Pinheiro2016-sharpmask}, and FastMask~\cite{Hu2017-fastmask}. 
We also compare to MCG~\cite{pont2017multiscale}, a top-performing non deep learning-based system~\cite{Hosang2015PAMI}. Since not relying on deep learning, MCG does not suffer from downsampling effects. 
\TODO{ Further comparison to instance segmentation systems is not conducted as those systems use class information for segmentation~\cite{he2017mask}.}

\subsection{Results on the LVIS dataset}
\label{sec:resLVIS}
Tab.~\ref{tab:resultsLVIS} shows the quantitative results on the LVIS dataset. The numbers reveal that our proposed system outperforms all other systems. Compared to AttentionMask using the same backbone network the AR@100 improves by $0.021$ with improvements across all scales of objects as AR$^S$@100, AR$^M$@100 and AR$^L$@100 indicate. Compared to AttentionMask with the more powerful ResNet-50 backbone and FastMask, the improvements in AR@100 are $0.017$ (AttentionMask) and $0.045$ (FastMask). Hence, our system is good enough to overcome the less powerful backbone. DeepMask and SharpMask both show very competitive results on large objects as indicated by the AR$^L$@100 results. This is mainly attributed to fewer downsampling steps in these systems that depend on an image pyramid. However, our proposed system still outperforms DeepMask and SharpMask on large objects and shows significant improvement in AR@100 across all scales of objects ($+0.059$ and $+0.052$). Interestingly, the non deep learning-based method MCG also produces very competitive results on AR$^L$@100 as no downsampling is applied. However, the results on medium and small object are much worse compared to our approach.


Qualitative results for the top-performing systems from Tab.~\ref{tab:resultsLVIS} are shown in Fig.~\ref{fig:qualRes}. It is clearly visible that across the different scenarios the results of our proposed system adhere very well to the object boundaries. For instance in the example in the first row, our result precisely captures fine details like the tail of the sheep, the visible front leg and the ears. In contrast, AttentionMask and FastMask generate rather blob like proposals containing a significant amount of background. The second row shows examples for very small objects with complex boundaries, which are still precisely segmented by our system. The results in the third and fourth row depict examples of humans in complex scenarios with spread legs and arms. Our proposals fit much better to the arms and legs capturing mostly the entire limbs. Note that the ground truth for the image in the forth row annotates the wet suits rather than the humans itself. The final row depicts an image with a highly cluttered environment where our system is still able to refine the proposals well especially around the corners of the objects. \TODO{Despite the strong overall performance, this example also shows a minor failure case for our system as the overlapping proposals for the nearby monitors indicate. Other typical sources of errors are usually related to the segmentation and occur in areas with low contrast or high texture.}

Further analyzing the accuracy of the segmentations, we use the typical segmentation metrics boundary recall (BR) and undersegmentation error (UE). To generate segmentations from annotations, we join per image the binary ground truth segmentations. For the results we join the binary segmentations of the best proposal per annotated object for each system. Thus, we create one proposal segmentation per image and system as well as one ground truth segmentation per image. The results on this segmentation-based evaluation are presented in Tab.~\ref{tab:resultsSeg}. Our proposed approach outperforms all deep learning based systems in terms of BR and UE. The significant improvement in terms of BR (between $39.5\%$ and $19.9\%$) indicates the stronger boundary adherence of our results. Therefore, our system segments more object boundaries significantly more precise. The results are also in line with the qualitative results in Fig.~\ref{fig:qualRes}. Compared to the non deep learning-based approach MCG, our results are on a similar level in terms of BR and better in terms of UE. Thus, we are able to combine the highly accurate segmentations of the non deep learning systems with the strong overall performance of the deep learning systems.

\begin{table}[t]
\renewcommand{\arraystretch}{1.2}
\centering
\caption{Comparison of the best proposals per image as segmentations using segmentation measures boundary recall (BR) and undersegmentation error (UE) on the LVIS dataset.}
\label{tab:resultsSeg}
{
\begin{tabular}{@{}lcccc@{}}
\hline
Method &  Backbone & BR$\uparrow$ & UE$\downarrow$ \\
\hline
MCG~\cite{pont2017multiscale} & - & 0.685  & 0.073 \\
\hline
DeepMask~\cite{Pinheiro2015-deepmask} & ResNet-50 & 0.488 & 0.087 \\
SharpMask~\cite{Pinheiro2016-sharpmask} & ResNet-50 & 0.561 & 0.080 \\
FastMask~\cite{Hu2017-fastmask} & ResNet-50 & 0.510  & 0.084 \\
AttentionMask~\cite{WilmsFrintropACCV2018} & ResNet-50  & 0.568 &  0.070  \\
\hline
AttentionMask~\cite{WilmsFrintropACCV2018} & ResNet-34  & 0.547 &  0.075 \\
Ours& ResNet-34 &  0.681  & 0.068  \\
\hline
\end{tabular}
}
\end{table}

\subsection{Ablation Studies}
In addition to the ablation studies in Sec.~\ref{sec:spxCalssifier} and Sec.~\ref{sec:intergration}, we show the influence of the number of superpixels, the segmentation method and the post-processing steps (Sec.~\ref{sec:intergration}) to further analyze our proposed refinement. Except for the post-processing, all ablation studies were conducted on a validation set of MS COCO dataset with a reduced model following~\cite{Hu2017-fastmask,WilmsFrintropACCV2018}. The reduced model contains only five scales and no post-processing to speed up training and testing.

\subsubsection{Influence of Segmentation}
\label{sec:ablInflSeg}
First, we evaluate the number of superpixels per scale with the used segmentation method FH~\cite{felzenszwalb2004efficient}. Compared to the setup presented in Sec.~\ref{sec:intergration} using approximately 8000 to 500 superpixels per scale, a reduction by the factor two does not improve the results (see Tab.~\ref{tab:ablInflSeg}). However, if we add ground truth edges to the segmentations a coarser segmentation is beneficial. This indicates that a strong oversegmentation is not helpful. Further increasing the number of superpixels is impossible due to memory constrains.

Similar conclusions can be drawn from the results of different segmentation methods. We compare the FH segmentation to the popular SLIC~\cite{achanta2012slic} method as well as ETPS~\cite{yao2015real}, ERS~\cite{liu2011entropy} and SEEDS~\cite{van2012seeds}: the top 3 methods in the superpixel benchmark of~\cite{stutz2018superpixels}. All four methods show inferior results compared to FH (see Tab.~\ref{tab:ablInflSeg}). This correlates with a higher oversegmentation error (up to $+17.6\%$) that those methods exhibit as they partition the image in rather regular superpixels of similar size. Thus, even highly uniform objects are artificially split into many pieces that have to be merged later on.

\begin{table}[t]
\renewcommand{\arraystretch}{1.2}
\centering
\caption{Results of different segmentation methods and numbers of superpixels used in the refinement approach. Numbers of superpixels are approximated across the validation dataset.}
\label{tab:ablInflSeg}
{
\begin{tabular}{@{}lcccc@{}}
\hline
Segmentation & \#Superpixels &  AR@10 & AR@100  & AR@1000 \\
\hline
FH~\cite{felzenszwalb2004efficient} & $8000-500$ & 0.160 &  0.296 & 0.382 \\
FH~\cite{felzenszwalb2004efficient} & $4000-250$ & 0.151 &  0.282 & 0.370 \\
\hline
FH~\cite{felzenszwalb2004efficient} with GT & $8000-500$ & 0.182 &  0.345 & 0.462 \\
FH~\cite{felzenszwalb2004efficient} with GT & $4000-250$ & 0.198 &  0.385 & 0.527 \\
\hline
ETPS~\cite{yao2015real} & $8000-500$ & 0.157 &  0.294 & 0.377 \\
ERS~\cite{liu2011entropy} & $8000-500$ & 0.143 &  0.278 & 0.368 \\
SEEDS~\cite{van2012seeds} & $8000-500$ & 0.148 &  0.269 & 0.344 \\
SLIC~\cite{achanta2012slic} & $8000-500$ & 0.137 &  0.267 & 0.356 \\
\hline
\end{tabular}
}
\end{table}

\begin{table}[t]
\renewcommand{\arraystretch}{1.2}
\centering
\caption{Results of the proposed refinement approach using different post-processing steps on the LVIS dataset.
}
\label{tab:ablPP}
{
\begin{tabular}{@{}lcccccc@{}}
\hline
Post-processing & AR@10 & AR@100  & AR@1000 \\
\hline
none & 0.076  & 0.184  & 0.276 \\
bilateral filtering & 0.079  & 0.191  & 0.288 \\
+ opening & 0.079 &  0.192 & 0.292 \\
+ closing & 0.079 &  0.193 & 0.293 \\
+ near duplicate removal & 0.092 &  0.206 & 0.290  \\
\hline
\end{tabular}
}
\end{table}

\subsubsection{Post-processing of Superpixel Results}
We further analyze the influence of the three step post-processing, described in Sec.~\ref{sec:intergration}. Tab.~\ref{tab:ablPP} shows the results without post-processing and step-wise added post-processing on the LVIS dataset. Overall, the results in terms of AR are increased by $5.1\%$ for AR@1000 and up to $21.1\%$ for AR@10 with a positive effect of each post-processing step. Most important are the bilateral filtering and the duplicate removal. Especially the duplicate removal shows a strong increase in terms of AR@10 ($+16.5\%$) compared to only applying the other two post-processing steps. This indicates a rather large number of duplicates or near duplicates in the original results due to the use of superpixels.

\section{Conclusion}
In this paper we address the problem of imprecise object segmentations due to downsampling in CNN-based object proposal generation systems. To tackle the problem, we presented a superpixel-based refinement approach on top of the state-of-the-art class-agnostic object proposal generation system AttentionMask. Our refinement utilizes highly accurate superpixel segmentations, superpixel pooling and a novel superpixel classifier to enhance the initial results. The experiments revealed improvements of up to $26.0\%$ for our overall system on the LVIS dataset against original AttentionMask with further improvements compared to other state-of-the-art systems. We also showed the superiority of our proposals in terms of segmentation measures boundary recall (max. $+39.5\%$) and undersegmentation error (max. $+23.0\%$) compared to state-of-the-art systems. In combination with the qualitative results, this indicates the strength of our proposals adhering to object boundaries. Ablation studies on the segmentation methods and the number of superpixels reveal the need for a highly accurate segmentation with low oversegmentation error, thus not artificially dividing uniform image areas.






\bibliographystyle{IEEEtran}
\bibliography{IEEEabrv,lit}

\end{document}